\documentclass[runningheads]{llncs}

\usepackage{graphicx}
\usepackage{amsmath}
\usepackage{amssymb}
\usepackage{booktabs}
\usepackage{url}
\usepackage{algorithm}
\usepackage{cite}
\usepackage{multirow}

\title{Bonsai: A Framework for Convolutional Neural Network\\
Acceleration Using Criterion-Based Pruning}

\titlerunning{Combine}

\author{
Joseph Bingham\inst{1,2} \and
Sam Helmich\inst{2}
}

\authorrunning{Bingham and Helmich}

\institute{
Rutgers University, New Brunswick NJ 08901, USA
\and
John Deere Intelligent Solutions Group, Urbandale, IA 50322, USA\\
\email{\{BinghamJosephC, HelmichSamR\}@JohnDeere.com}\\
\url{https://www.deere.com/en/our-company/john-deere-careers/work-here/isg/}
}

\begin{document}

\maketitle

As the need for more accurate and powerful Convolutional Neural Networks (CNNs) increases, so too does the size, execution time, memory footprint, and power consumption. To overcome this, solutions such as pruning have been proposed with their own metrics and methodologies, or criteria, for how weights should be removed. These solutions do not share a common implementation and are difficult to implement and compare. In this work, we introduce Combine, a criterion- based pruning solution and demonstrate that it is fast and effective framework for iterative pruning, demonstrate that criterion have differing effects on different models, create a standard language for comparing criterion functions, and propose a few novel criterion functions. We show the capacity of these criterion functions and the framework on VGG inspired models, pruning up to 79\% of filters while retaining or improving accuracy, and reducing the computations needed by the network by up to 68\%.

\section{Introduction}
\subsection{Background}

Convolutional Neural Networks [13] are a form of Neural Network that utilize convolutional filters, or sliding windows, to gain a deeper understanding about the data it is training over. By giving more importance to information based on the locality of the data, these models are incredibly powerful when applied to computer vision and image processing tasks [12]. Because of their strength within this crucial application, many research projects have focused on improving their accuracy [7], ability to be trained [5], robustness [1] and other useful qualities. As such, over the past few years, there has been a trend for these networks to grow larger and deeper, greatly increasing the number of parameters and operations [11]. This has presented a challenge for integrating state-of-the-art CNNs onto embedded systems, whose main constraints are memory, time and power consumption. Another system with these constraints are multi-threaded systems, such as web-based services which service multiple requests with large, but finite, resources. In these, and other applications, there arises the need for a way to reduce the size of a CNN without sacrificing its capacities and capabilities.
There has been work in the area of selecting the correct size and architecture for a CNN [10], but the fact still remains that deciding the correct hyperparameters for a model is still an open problem. Work in the field of knowledge distillation [2] has shown promise, but they make the initial training before distillation and retraining difficult as these operations still require a large amount compute power; thus they reduce the capacity to retrain models efficiently, especially on small embedded systems. These two methods constitute techniques for how to start with a sufficiently small model, and thus in order to resize of the model, pruning, or removing nodes [11], is required.

\subsection{Terminology}
In this paper, the term pruning is used to refer to any method that removes filters or nodes from a neural network. However, there are methods that used the nomenclature pruning, such as soft pruning [8] Soft pruning comprises techniques for better utilizing weights that do not contribute to the accuracy of a model.
Another set of methods that take on the name pruning is dynamic pruning [6]. These methods seek not to remove weights, but rather iteratively load only the portions of the full model that are needed for evaluating a particular input.
Although these methods certainly optimize the use of memory, as they are described, neither of these methods actually remove any nodes, and therefore do not remediate the issues within the scope of this paper. Therefore in this paper pruning will refer to what is often referred to in the literature as hard pruning [11], or any destructive method that accelerates the execution of a CNN.
Even restricting the scope to using only this definition of pruning, there still exists variation within the literature as how to implement such methods. The main differences within the proposed solutions manifest in two key ways. The first difference is how these methods determine which filters to prune. Any method used to make this decision is what is referred to as the pruning criterion. These differences will be explored in the related works section, and their exploration is the impetus for this work.
The second difference is when and in what order this criterion is applied in the pruning process. The prevailing techniques in previous works break into two categories. The first of these methods is a static application, which is hall- marked by the criterion being applied to all filters and then pruning occurs after. These methods are often faster as information is able to be memoized. This is in contrast to progressive methods, which will apply the criterion while it prunes, allowing for the evolution of information to be taken into account for the criterion. Note that although it may be less efficient to do so, any static method may be replicated by a progressive method, and is the first abstraction that our solution leverages.

\subsection{Contributions}

The goal of this work is to create a framework which will unify all appropriate approaches with further abstractions. This is done through Combine, a criterion-based pruning framework for the acceleration of Convolutional Neural Networks. This solution will have the following attributes:

\begin{itemize}
    \item \textbf{General:} Allow for any pruning technique to be approximated and use an existing machine learning library
    \item \textbf{Effective:} Faithfully prune nodes based on the given criterion for any CNN architecture
    \item \textbf{Efficient:} Prune nodes with minimal computation dedicated to pruning, avoiding full model evaluation when possible
\end{itemize}

The first two attributes of this solution will allow users to recreate previous pruning paradigms. This helps with not only the comparison of techniques for the correct selection of methods for a given architecture and domain, but also will help with reproducing results from previous work. Further, this framework will facilitate more development in the space of pruning.
All implementation of this framework were done with Keras [3] to allow for an easy integration with existing solutions3 Moreover, frameworks such as Keras aim to abstract away the more tedious and difficult aspects of their domain and allow for more plug-and-play interaction. This aspect is also adopted in Combine.
The final attribute comes from the use case. It is assumed that the user is computationally and time limited in their ability to execute the CNN. As such, it is unlikely that the users will have a large computational base to perform pruning. Although this restriction is not entirely necessary, it can be shown that Combine achieves this goal and outperforms more computationally expensive methods.
The contributions of this work are as follows:

\section{Related Works}
Existing solutions in the space of pruning take the form of a domain specific model with a proposed method for determining which of the filters need to be pruned [14][9]. These methods, although they often have promising results, lack any clear hierarchy as far as which would be best for a model that does not fit within their own domain or architecture. This is further complicated by the lack of uniformity in the language used by authors within this space.
Work such as [11] have proposed ways to prune CNNs using different criterion. However, instead of the functions mapping from the filter space to a value, Li et al. prunes based on feature maps. This requires that data be ran through the model and metadata be collected at each step of pruning. This is a costly operation, as the assumption is that running the model alone (i.e. without collection metadata) is an expensive operation.
Further, their implementation necessitates a hybrid approach between static and progressive, where they will establish their criterion values at the beginning of pruning a given layer, without updating any values in the process. Li et al. also establishes a good survey of the state of the art in pruning. While this work does outline the concepts of criteria well, it provides no analysis as to how one would apply them to any model to compare which would be best for a given architecture.
Similarly, the solution proposed by [4] also utilizes batch running images during an iterative pruning process, giving it an unforgiving complexity to be able to run on a limited architecture.

\section{Methods}
\subsection{Prunable Layers}

CNNs are made up of diverse layer types, but for the purposes of the Combine procedure, the layers are divided into two types - prunable and not prunable. Our primary work treats only dense and convolution layers as prunable, and other layers, such as pooling, normalization, or dropout layers are considered as non-prunable, but are important to the pruning process. Despite convolution and dense layers performing their functions in fundamentally different capacities, Combine relies on treating their weights in similar manners. The tensor describing the weights for the convolution layer is $(k_1 × k_2 × I × H )$, where $k_1 , k_2$ are the height and width of the convolution kernel, $I$ the number of input channels, and $H$ is the number of output channels, which corresponds to the number of convolution filters in the layer. For dense layers, the weights are represented by a $(I,H)$ tensor, where $I$ is the number of inputs and $H$ is the number of outputs, which corresponds to the number of hidden nodes in the layer. To Combine, a hidden node is equivalent to a convolution filter, and the term filter is used to describe them interchangeably within layers during the procedure. Even though these types of layers can be considered simultaneously, it is demonstrated in Section 4 that there are cases where pruning only one type of layer may be beneficial.

\subsection{Criterion Functions}
A criterion function is a measure of the relative contribution to the model of a filter in a particular layer. A criterion function can be any map of the form $\textit{f} : \mathbb{R}^{D×H} \longrightarrow \mathbb{R}^H$ , with the $h^{th}$ value corresponding to the criterion value of the $h^{th}$ filter. In the case of evaluating filters in a dense layer $D = I$, and in the case that the filter is in a convolution layer $D = k_1 × k_2 × I$. Higher output values of the criterion function should indicate that the filter is more valuable to the model, and each filter’s criterion value will be used to discriminate when choosing which filters to delete and which to keep. For simplicity, many criterion functions are specified as $\textit{f} : \mathbb{R}^D \longrightarrow \mathbb{R}$ and repeated on each filter. The criterion functions chosen here are reflective of that implementation: standard deviation, range, mean absolute value, and maximum absolute value. However, it is possible to specify criterion functions that then become layer-specific, such as a rank-ordering of criterion functions applied to individual layers.
Criterion functions can be applied in either a progressive or static fashion. In the static approach, the criterion function is applied to each layer before any pruning occurs, and in the progressive approach, the criterion function is applied after the previous layers were pruned. Static pruning is a faster approach, but the progressive approach aligns more closely with the idea that the criterion function is measuring the relative value to the model of the individual filter, since the contribution to the model necessarily changes as previous layers are pruned.

\subsection{Thresholds}
Once a criterion function has been selected, the final preparation is the selection of the threshold for pruning. Since higher criterion function values should indicate more usefulness to the network, any filter with a criterion value be- low the selected threshold will be discarded. Since it is possible to construct a criterion function to account for large differences in number and magnitude of weights (such as a rank-ordering or a percentile function), there only needs to be a single threshold defined for the entire pruning procedure.

\subsection{Combine}
To execute the algorithm, as shown in Algorithm 1, one must determine what types of layers to consider as prunable (either convolution, dense, or both), choose a criterion function, decide whether to apply the criterion function in a progressive or static manner, and choose a threshold. Then, proceeding through the model, the criterion values for filters are precalculated and memoised (in the static case) or calculated (in the progressive case), and any filters with criterion values below the selected threshold are fully removed from the model. The only exception is when a threshold is sufficiently large that would indicate that all filters should be removed. In this case, the filter with the largest criterion value is preserved as the only filter in the layer.
The Combine procedure is then a careful orchestration to ensure that the subsequent layers are not expecting information from output channels that no longer exist. While following layers may not be prunable, they may still need weights removed. In this respect, there are two type of layers - pass through layers that preserve the number of input channels (such as normalization or pooling layers), and layers that modify the fundamental shape of the feature map (such as convolution, dense, or flattening layers). Pass through layers can be easily modified by stripping out the portions of the feature map that correspond to the filters that were pruned in the previously pruned layer. In convolution and dense layers, the input channels that correspond to removed filter layers from the previously pruned layer need to be fully removed from the weight tensors for that layer. Flatten layers condense the feature map passing through to a single vector, and by tracking which elements post-flattening came from the removed filters in the previously pruned layers allows for the removal of those inputs to the next layer.

\begin{algorithm}[t]
\caption{Combine}
\label{alg:combine}
\textbf{Input:} Model $M$, Criterion $f$, Threshold $t$

\textbf{for} layer $l_i$ \textbf{in} $M$ \textbf{do}

\hspace{.5em} \textbf{if} $l_i$ is prunable \textbf{and} not last layer \textbf{then} 

\hspace{.75em} Initialize pruned filter index as empty

\hspace{.75em} \textbf{for} filter $m_j$ \textbf{in} $l_i$ \textbf{do}

\hspace{1em} \textbf{if} $f(m_j) < t$ \textbf{then}

\hspace{1.25em} Remove $m_j$

\hspace{1.25em} Append $j$ to list of pruned filter indices

\hspace{1em} \textbf{end if}

\hspace{.75em} \textbf{end for}

\hspace{.5em} \textbf{end if}

\hspace{.5em} \textbf{if} length(pruned filter index) $> 0$ \textbf{then}

\hspace{.75em} $k = 1$

\hspace{.75em} \textbf{while} $l_{i+k}$ is not prunable \textbf{do}

\hspace{1em} collect transformation metadata from $l_{i+k}$

\hspace{1em} $k += 1$

\hspace{.75em} \textbf{end while}

\hspace{.75em} remove input filters at indices from $l_{i+k}$ accounting for transformations

\hspace{.5em} \textbf{end if}

\textbf{end for}

\end{algorithm}

\subsection{Choosing a Threshold}
Choosing a threshold is a critical consideration in this procedure. Given a particular threshold, the pruning itself is very quick, and this can be leveraged to evaluate choices for the threshold value. In order to evaluate how well the pruning performed, a small representative version of a validation dataset is needed to evaluate the performance of the pruned model. As described in Algorithm 2, iterative pruning using different thresholds can be used to build out an estimation of the relationship between the percentage of the model pruned by a particular threshold and the resulting model accuracy (or other desirable loss function), called the threshold function.
In many cases, such as seen in Figure 1, there is a notable plateau in the threshold function, with different criterion having different plateau heights and lengths. It is advisable to choose the threshold that corresponds to a value near the plateau dropoff. If this does not achieve the required reduction in model size, the model can be retrained and then pruned again using the same procedure, or a lower threshold value can be chosen. It is important to recognize how much error is tolerated in the model, and should be taken into consideration when choosing a threshold.

\begin{algorithm}[t]
\caption{Iterative Pruning Threshold Function}
\label{alg:pruning}
\textbf{Input:} Model $M$, Criterion $f$, Threshold $t$, Loss Function $L(M, f, t)$

$t_0 = $ min($t$)

Prune $M$ w.r.t $f, t_1$

$l_1 = L(M, f, t_1)$

$t_2 = $ max($t$)

Prune $M$ w.r.t $f, t_2$ 

$l_2 = L(M, f, t_2)$

Let $l^*$ be the sorted $l$ values and $t^*$ be the threshold values sorted by their corresponding loss

$i = 2$

\textbf{while} max($l_i^* - l_{i+1}^* <$ largest desirable gap \textbf{do}

\hspace{.5em} $i += 1$

\hspace{.5em} Find $j | l_j^* - l_{j+1}^* = $ max($l_i^* - l_{i+1}^*$)$\forall i$ 

\hspace{.5em} Prune $M$ w.r.t $f, t_i = \frac{t_j^* - t_{j+1}^*}{2}$

\hspace{.5em} $l_i = L(M,f,t_i)$

\textbf{end while}

\end{algorithm}
\subsection{Choosing a Criterion Function}
The threshold function can be utilized to choose the best criterion function. The area under the curve of the threshold function can be used as a metric for comparison for criterion functions applied on the same model. This metric should not be used to compare criterion function across models, as the area under the curve will be dependent on the accuracy of the initial model, and therefore the range of values will be model dependent.

\begin{figure}[tp]
  \centering
  \includegraphics[height=.75\textwidth]{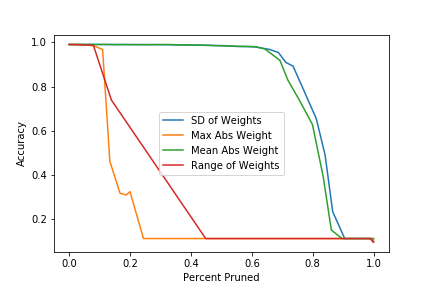}
  \caption{Criterion performance in Combine procedure on the Balanced network trained on MNIST Data
}
  \label{fig:balanced}
\end{figure}

\section{Results}

Since the Combine method works for both dense and convolution layers, performance was assessed on three types of models: a model that has many more convolution weights than dense weights (Model A), a model with balanced convolutional and dense weights (Model B), and a model that has more dense weights than convolutional (Model C). The model architectures chosen are VGG- inspired, as seen in Figure 3.

\begin{table}[h]
    \centering
    \small
    \caption{Model top-1 accuracy of various models on CIFAR-10 and MNIST}
    \begin{tabular}{lll}
        Model & MNIST & CIFAR-10 \\
        \hline \hline
        Convolutional Heavy (A) & 99.04\% & 77.14\% \\ 
        \hline
        Balanced (B) & 99.07\% & 73.26\% \\ 
        \hline
        Dense Heavy (C) & 98.00\% & 49.72\% \\
        \hline
    \end{tabular}
\label{tab:my_label}
\end{table}

Each of the 3 models were trained on both MNIST and CIFAR-10 datasets, with the pre-pruned model top-1 accuracy documented in Table 1.

\begin{figure}[tp]
  \centering
  \includegraphics[height=.75\textwidth]{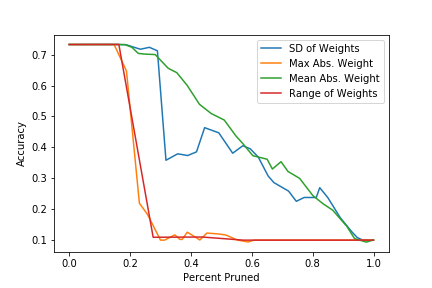}
  \caption{Criterion performance in Combine procedure on the Balanced network trained on CIFAR-10 Data
}
  \label{fig:balanced_cifar}
\end{figure}

\begin{figure}[tp]
  \centering
  \includegraphics[height=.75\textwidth]{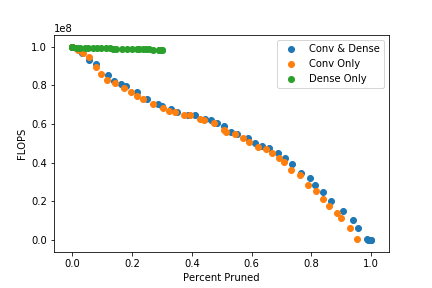}
  \caption{Computation requirement improvements on Balanced model on MNIST data with pruning only dense layers, convolution layers, or both.}
  \label{fig:layers}
\end{figure}

\subsection{Accuracy}
The size and style of these architectures is larger than needed to achieve excellent performance on MNIST, and smaller than needed to achieve top performance on the CIFAR-10 data, but this was an intentional choice. This should lead to pruning most of the MNIST trained models, and pruning a much smaller amount of the CIFAR-10 trained models.
There are two major factors under consideration - which layers should be considered as prunable (dense, convolutional, or both), and which criterion function should be applied to the model (standard deviation, maximum absolute value, mean absolute value, or range). For each combination of layer selection and model, all four criterion functions were applied to get each of the threshold functions. The threshold functions for Model B (Balanced) trained on MNIST data pruning both layer types can be found in Figure 1 and the threshold functions for the same model trained on CIFAR data pruning both layer types can be found in Figure 2. For each of the threshold functions, the area under the curve was calculated and used to determine the best criterion function. In Figure 1, all four criterion functions are plotted with the longest plateau (standard deviation of filter weights) has an area under the curve of 0.824, while the worst (maxi- mum absolute value) has an area under the curve of 0.224. This procedure was repeated for all combinations of model and layer selection, with the results for the models trained on MNIST recorded in Table 2 and the CIFAR-10 trained model’s performance recorded in Table 3.
In general, it was found that applying the Combine procedure while treating both dense and convolution layers as prunable provides for the largest area under the curve (and therefore the best performance) for most criterion functions, with the exception being the dense weight heavy model trained on the CIFAR-10 data. This is largely due to the small number of convolution layers, and each one having a critical contribution to the model.
After identifying the best criterion function and selection of prunable layers, each model was pruned and retrained on the same training data it was originally trained on. As seen in Table 4 there are two cases where the post-pruning ac- curacy was minimally smaller than the original model, at most a loss of 0.34\%, and the other case the post-pruning accuracy increased by 0.29\% reflecting that the pruning was able to identify filters that were either under-contributing or contributing noise and allowed retraining to further enhance the remaining useful filters. As expected, Bansai was able to prune most of the MNIST trained models - in one case (Model A - Convolution Heavy on MNIST), Bansai was able to prune 79\% of the model while improving overall accuracy, as found Table 5. On the CIFAR-10 models, the Combine procedure was able to prune large chunks of the model, 16\%-38\%, while increasing accuracy, demonstrating that models can be improved and accelerated by the Combine procedure, even if they were relatively small to begin with.

\begin{figure}[tp]
  \centering
  \includegraphics[height=.75\textwidth]{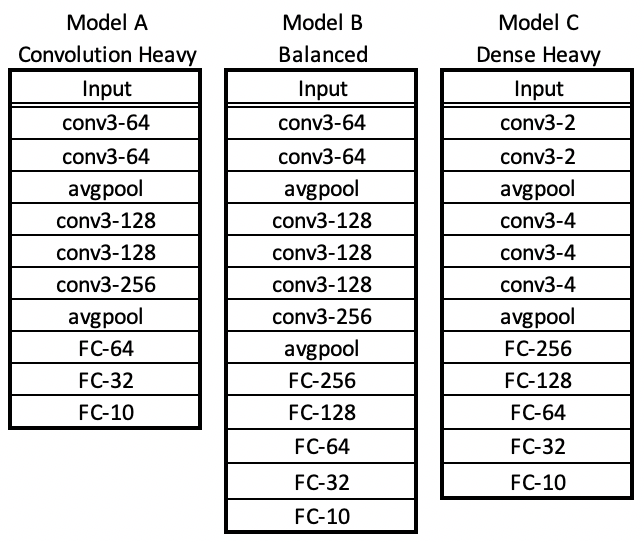}
  \caption{VGG inspired model architectures}
  \label{fig:arch}
\end{figure}

\begin{table}[h]
    \centering
    \small
    \caption{Combine Performance across criterion frunction on MNIST trained models. * denotes best criterion function for the model and layer selection, and \textbf{bold} denotes the best criterion function and layer selection for the model.}
\begin{tabular}{llllll}
\hline
Pruned                 & Model & SD     & Max Abs & Mean Abs & Abs Range \\ \hline \hline
\multirow{3}{*}{Both}  & A     & 0.860* & 0.245   & 0.855    & 0.219     \\ \cline{2-6} 
                       & B     & \textbf{0.824*} & 0.244    & 0.805    & 0.324     \\ \cline{2-6} 
                       & C     & \textbf{0.806*} & 0.781   & 0.776    & 0.778     \\ \hline
\multirow{3}{*}{Conv}  & A     & \textbf{0.879*} & 0.232   & 0.870    & 0.230     \\ \cline{2-6} 
                       & B     & 0.801  & 0.230   & 0.789    & 0.182     \\ \cline{2-6} 
                       & C     & 0.045  & 0.036   & 0.061*   & 0.042     \\ \hline
\multirow{3}{*}{Dense} & A     & 0.174  & 0.169   & 0.179*   & 0.175     \\ \cline{2-6} 
                       & B     & 0.280  & 0.242   & 0.283*   & 0.235     \\ \cline{2-6} 
                       & C     & 0.804* & 0.780   & 0.777    & 0.779     \\ \hline
\end{tabular}
\end{table}

\begin{table}[h]
    \centering
    \small
    \caption{Combine Performance across criterion frunction on CIFAR-10 trained models. * denotes best criterion function for the model and layer selection, and \textbf{bold} denotes the best criterion function and layer selection for the model.}
\begin{tabular}{llllll}
\hline
Pruned                 & Model & SD     & Max Abs & Mean Abs & Abs Range \\ \hline \hline
\multirow{3}{*}{Both}  & A     & 0.293 & 0.192   & \textbf{0.343*}    & 0.177  \\ \cline{2-6} 
                       & B     & 0.431 & 0.239    & \textbf{0.474*}    & 0.242 \\ \cline{2-6} 
                       & C     & 0.166 & 0.155   & 0.167*    & 0.150     \\ \hline
\multirow{3}{*}{Conv}  & A     & 0.225 & 0.156   & 0.275*    & 0.158     \\ \cline{2-6} 
                       & B     & 0.363  & 0.185   & 0.425*    & 0.225     \\ \cline{2-6} 
                       & C     & 0.044  & 0.031*   & 0.043   & 0.038     \\ \hline
\multirow{3}{*}{Dense} & A     & 0.191  & 0.085   & 0.202*   & 0.081     \\ \cline{2-6} 
                       & B     & 0.280  & 0.131   & 0.311*   & 0.135     \\ \cline{2-6} 
                       & C     & 0.297* & \textbf{0.333}*   & 0.321    & 0.313     \\ \hline
\end{tabular}
\end{table}

\begin{table}[h]
    \centering
    \small
    \caption{Combine top-1 accuracy across models trained on MNIST after selecting a criterion function, pruning, and retraining the pruned model}
    \begin{tabular}{llll}
        Model & Original & Post-prune/retrain & \% Pruned \\
        \hline \hline
        A & 99.04\% & 99.33\% & 79.00\% \\ 
        \hline
        B & 99.07\% & 98.73\% & 76.52\% \\ 
        \hline
        C & 98.00\% & 97.90\% & 70.83\% \\
        \hline
    \end{tabular}
\label{tab:my_label}
\end{table}

\begin{table}[h]
    \centering
    \small
    \caption{Combine top-1 accuracy across models trained on CIFAR-10 after selecting a criterion function, pruning, and retraining the pruned model}
    \begin{tabular}{llll}
        Model & Original & Post-prune/retrain & \% Pruned \\
        \hline \hline
        A & 77.14\% & 78.36\% & 16.55\% \\ 
        \hline
        B & 73.26\% & 79.11\% & 38.76\% \\ 
        \hline
        C & 49.72\% & 50.32\% & 35.72\% \\
        \hline
    \end{tabular}
\label{tab:my_label}
\end{table}

\subsection{Performance}
Pruning convolution and dense layers have different effects on overall model compute requirements, due to how the filters are applied to the feature vectors. In a dense layer there is a single weight for each of the I input vectors and H output filters. The Floating Point Operations (FLOPs) needed for each dense layer is given by $2IH$. Since a convolution layer has to be convolved over the whole feature map (with height $r_h$ and width $r_w$ ) then the FLOPs needed for each convolution layer is $2k_1k_2r_hr_wIH$. Then removing a single dense filter results in a FLOPs decrease of $2I$ where removing a single convolution filter results in a FLOPs decrease of $2k_1k_2r_hr_wI$, meaning the performance improvement is also affected by the size of the convolution filter ($k1$ and $k2$) as well as the image size.
In the Combine procedure, choosing which layers to be considered as prunable has a profound impact on final model compute costs. By ensuring that convoluton layers can be successfully considered for pruning, the computational cost for the model is considerably larger than by only considering dense layers, as demonstrated in Figure 4. When it is feasible to consider pruning both convolution and dense layers, the maximum acceleration can be achieved, but in cases such as Model C on the CIFAR-10 data, there may not be sufficient convolution layers to effectively consider them prunable.

\begin{table}[h]
    \centering
    \small
    \caption{Combine Performance across models trained on MNIST after selecting a criterion function, pruning, and retraining the pruned model}
    \begin{tabular}{llll}
        Model & Original & Post-prune & \% Reduction \\
        \hline \hline
        A & 105,595,549 & 32,973,437 & 68.77\% \\ 
        \hline
        B & 107,757,094 & 37,178,977 & 65.50\% \\ 
        \hline
        C & 752,919 & 330,418 & 56.12\% \\
        \hline
    \end{tabular}
\label{tab:my_label}
\end{table}

\begin{table}[h]
    \centering
    \small
    \caption{Combine Performance across models trained on CIFAR-10 after selecting a criterion function, pruning, and retraining the pruned model}
    \begin{tabular}{llll}
        Model & Original & Post-prune & \% Reduction \\
        \hline \hline
        A & 158,355,901 & 120,584,938 & 23.84\% \\ 
        \hline
        B & 166,591,782 & 98,209,450 & 41.05\% \\ 
        \hline
        C & 953,727 & 772,011 & 19.05\% \\
        \hline
    \end{tabular}
\label{tab:my_label}
\end{table}

After following the procedure as laid out in Section 4.1 of choosing a criterion function and effective threshold and then re-training the model, a considerable model speedup can be demonstrated. For the MNIST trained models the accuracy (Table 4) drop was at most 0.66\% or less and the FLOPs speedup (Table 6) was between 56.12\% and 68.77\%, and on the CIFAR trained models the accuracy (Table 5) went up in every model while the FLOPs speedup (Table 7) was between 19.05\% and 41.05\%. If a larger accuracy drop is acceptable, there is still opportunity left on the table to further decrease the compute requirements for the models by either increasing the chosen threshold or repeating the pruning and retraining process.

\section{Discussion}
The Combine framework is valuable as it begets a tool implementation that is approachable to any model architect. By having useful abstractions for criterion functions, thresholds, prunable layers, and how the criterion function is applied, these items can become interchangeable and be easily applied to any model building effort.
Pruning should be a part of the model building process with criterion function selection being parallel to choosing an optimizer or an activation function, and the selection of which layers are prunable should be a conscious effort after choosing the number of filters or hidden nodes to have in each layer. Having this framework implemented in common tools will also allow existing models to be easily pruned and placed back in production.
In a world where computer vision tasks are becoming a fundamental back- bone of automation, it is important to recognize that there are many use cases where there will not be sufficient connectivity or bandwidth to enable computer vision tasks to be executed on the cloud, and areas where putting large amounts of compute on edge systems is not economically feasible. Additionally, having the ability to take existing models and compress the model size and compute necessary to execute will allow speedups and improvements without replacing existing systems and software.
The amount of compute needed to execute models will also be a limiting factor in many applications where framerate is a critical consideration for the entire automation system. If models have compute requirements that are too high, edge compute may not be able to keep up with operational requirements of the model, acting as a restriction on the system. In these situations, pruning the existing model can speed up the entire system without changing the hardware or creating a new model.

\section{Conclusion}
There have been a variety of implementations of pruning for CNNs which can be united with a single framework. The Combine framework provides an approach- able mechanism for pruning new and existing neural networks by creating useful abstractions for criterion functions, thresholds, and how they are applied to the network. The framework demonstrated up a 68\% speedup on MNIST trained models and up to a 41\% speedup on CIFAR-10 based models with either negligible loss in accuracy or accuracy improvement, upon retraining. This has been demonstrated to provide considerable reductions in Floating Point Operations (FLOP) while largely preserving the accuracy of the model at hand. These techniques paired with the Combine framework make pruning approachable to an extent that it can be considered a core element of the model building process.

\bibliographystyle{splncs04}

\end{document}